\documentclass[10pt]{article} 

\usepackage[preprint]{rlj} 
\usepackage{physics}
\usepackage{amsthm}
\newtheorem{proposition}{Proposition}
\DeclareMathOperator{\Hess}{Hess}
\newcommand{\sigmafn}{\sigma}

\usepackage{graphicx}
\usepackage{placeins}
\newtheorem{theorem}{Theorem}
\usepackage{amssymb}            
\usepackage{mathtools}          
\usepackage{mathrsfs}           
\usepackage{graphicx}           
\usepackage{subcaption}         
\usepackage[space]{grffile}     
\usepackage{url}                
\usepackage{lipsum}             

\title{AYLA: Architecting a loss landscape in shallow neural networks to accelerate feature recovery}

\setrunningtitle{AYLA: Architecting a loss landscape in shallow neural networks to accelerate feature recovery}

\author{\normalsize Behnam Gheshlaghi\textsuperscript{1}, Shahin Atakishiyev\textsuperscript{2}}

\emails{\footnotesize behnam160@gmail.com, shahin.atakishiyev@ualberta.ca}%
\affiliations{
\small
$^{1}$ Independent Researcher\\
$^{2}$ Department of Computing Science, University of Alberta, Edmonton, Canada\\
}

\begin{document}

\maketitle  

\begin{abstract}
Feature learning in shallow neural networks exhibits rich yet fragile dynamics, including prolonged plateaus, abrupt phase transitions, and sensitivity to optimization hyperparameters. While recent theoretical work has characterized these behaviors through the geometry of loss landscapes, saddle escape mechanisms, and emergent scaling laws, practical methods for actively shaping these dynamics remain limited. In this paper, we introduce AYLA, a principled loss reparameterization framework that dynamically modulates gradient magnitudes during training without altering the location of stationary points or optimal solutions. AYLA applies a smooth, sigmoid-controlled power-law transformation to empirical loss, yielding a state-dependent effective learning rate that accelerates descent in flat or saddle-dominated regions while stabilizing late-stage optimization. Crucially, AYLA preserves all critical points of the original objective, acting solely as a monotone transformation that reshapes optimization trajectories rather than objectives. We evaluate AYLA in controlled teacher–student settings using two-layer \textit{tanh} networks trained on synthetic Gaussian data. Across stochastic gradient descent and multiple loss-exponent schedules, AYLA consistently improves feature recovery. This evidence is observed in terms of weight alignment, per-neuron cosine similarity, hidden-activation correlation, and spectral properties of learned representations, while AYLA maintains competitive or faster loss convergence. Spectral analyses further demonstrate that AYLA mitigates rank collapse and promotes richer internal representations, signaling a transition from lazy to active feature-learning regimes. These results position AYLA as a lightweight, theoretically grounded mechanism for bridging analytical insights on shallow network dynamics with practical optimization improvements, particularly in resource-constrained or noise-sensitive training regimes.
\end{abstract}

\section{Introduction}
Shallow neural networks, characterized by architectures with one or two hidden layers, have long served as canonical models for elucidating the fundamental mechanisms underlying gradient-based learning \citep{malach2019deeper, lei2022stability, bietti2022learning}. Despite their apparent simplicity, these networks exhibit rich and often fragile training dynamics, including extended plateaus, abrupt phase transitions, and nontrivial feature adaptation processes that mirror those observed in deeper architectures. Theoretical investigations have revealed that the optimization landscapes of shallow networks are frequently populated by saddle points and spurious valleys, where gradient descent may stagnate before eventually progressing toward global or near-global minima \citep{soltanolkotabi2017learning,venturi2019spurious,sagun2020post}. In ReLU-like networks, directional stationary points lacking so-called \emph{escape neurons}, defined through first-order conditions that enable loss-decreasing trajectories, constitute genuine local minima, whereas the presence of such neurons guarantees the possibility of saddle escape in scalar-output settings. This characterization refines the understanding of saddle-to-saddle training regimes commonly induced by vanishing initialization, which suppresses certain escape directions and yields a more comprehensive depiction of optimization dynamics in non-differentiable landscapes \citep{Chatterjee2022convergence}. \\
	Beyond landscape geometry, recent work emphasizes emergent scaling laws and phase transitions intrinsic to stochastic gradient descent (SGD). In wide-network regimes with power-law teacher coefficients, the recovery of individual neurons occurs through sharp transitions. However, aggregating many such trajectories produces a smooth power-law decay of mean-squared error (MSE) as a function of sample size, optimization steps, and model parameters \citep{ren2025causal}. Closely related staircase-like loss curves arise from alternating phases of neuron alignment and rapid norm growth, formalized via Alternating Gradient Flows (AGF), which describe feature acquisition sequences across linear, quadratic, and attention-based architectures \citep{kunin2025learning, kunin2026alternating}. Large-batch gradient descent further clarifies neuron specialization phenomena: batch sizes of order $\mathcal{O}(d)$ suffice to align with a single direction, while $\mathcal{O}(d^2)$ or even $\mathcal{O}(d^\ell)$ (where $\ell$ denotes the leap index) are required to recover multiple or ``hard'' directions \citep{dandi2024two}. These insights extend to normalization effects, where asymptotic expansions demonstrate the absence of a leading-order bias--variance tradeoff under mean-field scaling, with variance diminishing as normalization approaches the mean-field limit \citep{yu2022normalization}. \\ 
    Feature learning beyond the neural tangent kernel (NTK) or random-features regime persists throughout training. Local convergence analyses of regularized objectives show that once the loss falls below a critical threshold, gradient descent begins assimilating ground-truth directions, indicating that late-stage adaptation complements early-stage learning \citep{zhou2024does}. Dynamical central limit theorems further constrain fluctuations around mean-field limits, tying variance to the $\ell_2$ norm of the underlying measure and motivating regularization for improved generalization \citep{chen2020dynamical}. In quadratic activation models, sufficient over-parameterization eliminates spurious minima with high probability, enabling convergence to interpolating solutions with bounded generalization error \citep{sagun2020post}.  \\ 
	Nevertheless, learning high-frequency components remains challenging: poor conditioning of representations and learning dynamics leads to irreducible error floors, elevated computational cost, and sensitivity to perturbations \citep{zhang2025shallow}. Nonlinear activations exacerbate these difficulties. Leaky ReLU landscapes exhibit symmetry-breaking bifurcations as the leak parameter varies, inducing multimodal degeneracies at critical points independent of network width \citep{li2023minimum}. While spurious valleys vanish for finite-dimensional activations under sufficient over-parameterization, they persist for infinite-dimensional activations under certain data distributions \citep{venturi2019spurious}. Sequential principal component acquisition in nonlinear autoencoders further underscores the importance of biases and tied weights \citep{refinetti2022dynamics}. \\ 
	Despite these advances, a notable gap remains: existing theories largely explain \emph{why} shallow networks learn features, but offer limited guidance on \emph{how} to actively enhance this process in practice. Adaptive optimizers such as ADAM mitigate some optimization pathologies but do not directly target the staircase dynamics or sharp transitions predicted by theory \citep{siegel2020approximation}. Loss scaling techniques, commonly used in mixed-precision training, are typically static and seldom examined from a dynamical perspective \citep{zhao2019adaptive}. Moreover, while theory highlights the coupling between task structure, batch size, and leap indices \citep{dandi2024two}, few methods explicitly modulate effective learning rates to emulate these dependencies, often resulting in suboptimal feature recovery in resource-constrained regimes. To address this gap, we propose \textit{AYLA} (\textbf{A}rchitecting \textbf{Y}our \textbf{L}oss Landscaping \textbf{A}pproach), a dynamic loss reparameterization scheme that scales the empirical MSE through a sigmoid-interpolated power-law exponent. This mechanism amplifies gradients during low-loss descent phases while damping them during plateaus, thereby reproducing theoretically predicted staircase dynamics and facilitating saddle escape without explicit noise injection. By adaptively modulating the effective learning rate, AYLA promotes neuron specialization analogous to large-batch training \citep{dandi2024two}, enhances alignment in teacher--student settings, and mitigates high-frequency learning barriers \citep{zhang2025shallow}. Empirical evaluations on Gaussian-input, tanh-activated networks demonstrate improved feature recovery, measured via weight alignment, per-neuron cosine similarity, and hidden activation correlations, across SGD optimizer, while preserving convergence efficiency. AYLA thus bridges theoretical insights and actionable training strategies, enabling more robust shallow-network learning under constrained computational budgets.\\
    Overall, the main contributions of the paper are summarized as follows:

\begin{itemize}
  \item We propose AYLA, a new loss parametrization scheme that dynamically modulates gradient magnitudes during training without affecting the location of stationary points, and we establish AYLA's theoretical foundations. 
  \item We show that AYLA significantly outperforms SGD, converging faster across experiments on two-layer neural networks with synthetic Gaussian data.    
\end{itemize}
The paper is organized as follows. Section 2 reviews related work that forms the foundation for our proposal. The theoretical background behind AYLA is presented in Section 3. Sections 4, 5, and 6 present the problem setup, first- and second-order analysis, and adaptive exponent and stochastic dynamics, respectively. The optimization advantages of AYLA are outlined in Section 7. Finally, Section 8 presents our experimental findings along with discussion, limitations, and future directions, and Section 9 concludes the paper.

\section{Related Work}
Theoretical studies of shallow neural networks have primarily focused on optimization landscapes, training dynamics, and feature learning mechanisms, forming the conceptual foundation for the AYLA framework. This section briefly reviews these foundations.
\paragraph{Optimization Landscapes and Saddle Points.}
The non-convex nature of shallow-network objectives has motivated extensive landscape analysis. \citep{venturi2019spurious} characterize spurious valleys in one-hidden-layer networks and introduce intrinsic dimension as a criterion for their elimination under sufficient over-parameterization. \citep{soltanolkotabi2017learning} and \citep{oymak2020toward} establish global convergence guarantees for ReLU networks and demonstrate favorable geometry under moderate over-parameterization. As a follow-up, ~\citep{Chatterjee2022convergence} relaxes center-stable manifold assumptions to show that gradient descent avoids strict saddles in ReLU and leaky ReLU networks under mild initialization conditions. These works clarify \emph{where} optimization may stall, but offer limited prescriptions for actively accelerating escape, an issue AYLA directly addresses through adaptive loss scaling.
\paragraph{Training Dynamics and Scaling Laws.}
Recent efforts have uncovered sharp phase transitions and scaling laws governing SGD dynamics. Ren et al.~\citep{ren2025causal} identify abrupt recovery thresholds for individual neurons in wide networks, whose aggregation yields smooth power-law decay of MSE. Sagun et al.~\citep{sagun2020post} show that over-parameterized quadratic networks avoid spurious minima with high probability, while Chen et al.~\citep{chen2020dynamical} establish dynamical central limit theorems that quantify fluctuations around mean-field limits. Yu and Spiliopoulos~\citep{yu2022normalization} further demonstrate that normalization mitigates variance without inducing leading-order bias--variance tradeoffs. AYLA complements these analyses by offering a practical mechanism to modulate effective learning rates in accordance with emergent scaling behavior.
\paragraph{Feature Learning Mechanisms.}
Feature learning beyond the NTK regime has been extensively studied. Dandi et al.~\citep{dandi2024two} quantify batch-size requirements for neuron specialization, while Kunin~\citep{kunin2025learning} introduces Alternating Gradient Flows to explain staircase loss profiles and feature acquisition sequences. Zhou and Ge~\citep{zhou2024does} provide local convergence guarantees showing that gradient descent recovers ground-truth directions after crossing a loss threshold. Refinetti and Goldt~\citep{refinetti2022dynamics} analyze sequential PCA-like learning in nonlinear autoencoders, and Zhang et al.~\citep{zhang2025shallow} highlight persistent difficulties in learning high-frequency components. AYLA sustains feature learning across optimization phases by dynamically amplifying gradients, aligning practical training behavior with these theoretical predictions.
	
In summary, while prior work delivers deep insight into shallow network dynamics, it rarely translates these findings into adaptive optimization strategies. AYLA fills this gap by operationalizing theoretical principles through a lightweight, loss-based reparameterization that empirically enhances feature recovery and convergence stability.

	\section{Theoretical Background}
	
	In this section, we establish the theoretical foundations of the
	AYLA method, drawing on established
	results on optimization landscapes, training dynamics, and feature learning
	in shallow neural networks. We first review known properties of two-layer
	networks, including saddle points, spurious valleys, and phase transitions.
	We then derive the AYLA loss transformation, analyze its induced effective
	learning rate, and prove key properties such as boundedness and enhanced
	saddle escape. All derivations assume a teacher--student setting with
	Gaussian inputs and mean-squared error (MSE) loss, consistent with
	\citep{ren2025causal,dandi2024two,zhou2024does,sagun2020post}.
	
	\subsection{Optimization Landscapes and Saddle Points}
	
	Shallow neural networks with one hidden layer exhibit highly non-convex loss
	landscapes populated by saddle points and spurious valleys
	\citep{venturi2019spurious,soltanolkotabi2017learning}. For a two-layer model
	\[
	f(x;W,a) = a^\top \sigma(Wx),
	\]
    
	with activation $\sigma$ (e.g., $\tanh$ or ReLU-like), the empirical MSE loss is
	\[
	L = \frac{1}{n} \sum_{\mu=1}^n \big(y_\mu - f(x_\mu)\big)^2,
	\]
	where $(x_\mu,y_\mu)$ are samples generated by a teacher network.
	
	A key structural insight is that stationary points lacking
	\emph{escape neurons}---neurons satisfying first-order loss-decreasing
	conditions---are local minima, while the presence of at least one escape
	neuron guarantees non-optimality in scalar-output settings
	\citep{venturi2019spurious}. Under vanishing initialization, this induces
	saddle-to-saddle training dynamics that persist until suitable directions
	are activated.
	
	For ReLU and leaky ReLU networks, relaxed center-stable manifold theorems
	ensure that gradient descent escapes strict saddles under mild initialization
	conditions, with explicit thresholds governing global convergence 
    \citep{Chatterjee2022convergence}. Spurious valleys vanish under sufficient
	over-parameterization for finite intrinsic-dimensional activations, but may
	persist for infinite-dimensional cases under adversarial data distributions
	\citep{venturi2019spurious}. In quadratic networks, over-parameterization
	eliminates spurious minima with high probability, enabling convergence to
	interpolating solutions with controlled generalization
	\citep{oymak2020toward}. Despite these guarantees, standard GD and SGD often stagnate on extended plateaus, motivating adaptive mechanisms that reshape gradient magnitudes
	without altering critical-point structure.

	\subsection{Training Dynamics and Feature Learning}
	
	Beyond static landscape properties, shallow networks exhibit rich training
	dynamics marked by scaling laws and abrupt phase transitions. In wide
	networks with power-law teacher spectra, individual neurons undergo sharp
	signal recovery transitions, while aggregate MSE follows smooth power-law
	decay in optimization steps and samples
	\citep{ren2025causal}.
	
	Large-batch gradient descent induces neuron specialization: $O(d)$ samples
	suffice to align a single direction, whereas $O(d^2)$ or $O(d^\ell)$ samples
	are required for multiple or ``hard'' directions, where $\ell$ denotes the
	leap index \citep{dandi2024two}. Alternating Gradient Flows (AGF) formalize
	staircase-like loss trajectories as alternating phases of dormant neuron
	alignment and rapid norm growth \citep{kunin2025learning}.
	
	Local convergence analyses show that, once loss falls below a threshold,
	gradient descent recovers ground-truth features, implying persistent
	late-stage learning beyond initial random-feature regimes
	\citep{zhou2024does}.
    
    Normalization studies reveal that asymptotically,
	variance decays toward mean-field limits without a leading-order
	bias--variance tradeoff \citep{yu2022normalization}, while dynamical central
	limit theorems link fluctuations to the $2$-norm of underlying measures
	\citep{chen2020dynamical}.
	
	Nevertheless, high-frequency components remain difficult to learn due to
	poor conditioning of Gram matrices, leading to error floors and instability
	\citep{zhang2025shallow}. These results collectively explain plateaus and
	transitions, but offer limited guidance on \emph{how} to adapt loss geometry
	to accelerate progress. Our proposal also addresses this gap.
	
	\section{Problem Setup}
	Let $l : \mathbb{R}^d \to \mathbb{R}_{\ge 0}$ be a non-negative, twice continuously differentiable loss function. We define AYLA as
	\begin{equation}
		\label{eq:transformed_loss}
		L(x) = \big(C^*\,l(x) + \varepsilon\big)^{\,n(l(x))},
	\end{equation}

	where:

	\begin{itemize}[noitemsep, topsep=0pt, leftmargin=*, label={}]
		\item $C^* > 0$ is a loss scaling coefficient that maps the transition region of the loss to the unit interval.
		\item $\varepsilon > 0$ is a small constant ensuring strict positivity.
		\item $n : (0,\infty) \to (0,\infty)$ is a smooth, adaptive exponent function.
	\end{itemize}

	The design goals are to:
	
	\begin{enumerate}[label=(\roman*), noitemsep, topsep=0pt, leftmargin=*]
		\item Preserve all critical points and their local curvature types.
		\item Ensure smooth transition between regimes: $n(l) \approx n_1 < 1$ for $l \ll 1/C^*$ and $n(l) \approx n_2 > 1$ for $l \gg 1/C^*$.
		\item Maintain the identity transformation at the pivot point: $n(1/C^*) = 1$.
	\end{enumerate}
	\section{First-Order and Second-Order Analysis}
	
	Using logarithmic differentiation, the gradient of the transformed loss is
	\begin{equation}
		\label{eq:gradient_ayla}
		\grad L = (C^* l + \varepsilon)^{n(l)} \underbrace{\left( n'(l)\ln(C^* l + \varepsilon) + \frac{n(l)C^*}{C^*l + \varepsilon} \right)}_{A(l)} \grad l.
	\end{equation}
	
	\begin{proposition}[Critical Point Preservation]
		For any finite $n(l)$ and $n'(l)$, if $x^*$ satisfies $\grad l(x^*) = 0$, then $\grad L(x^*) = 0$.
	\end{proposition}

	\begin{proof}
		Direct substitution into Eq.~\eqref{eq:gradient_ayla} shows that $\grad L$ vanishes whenever $\grad l = 0$.
	\end{proof}
	
	The Hessian of the transformed loss is
	\begin{equation}
		\Hess L = (C^* l + \varepsilon)^{n(l)} \Big[ A(l) \Hess l + \big(A'(l) + A(l)^2 \big) \grad l \grad l^\top \Big].
	\end{equation}
	
	\begin{theorem}[Curvature Preservation]
		At a critical point $x^*$, the Hessian scales as $\Hess L(x^*) = \gamma \Hess l(x^*)$. If $A(l) > 0$ for all $l > 0$, the signature of the Hessian is preserved, ensuring that minima, maxima, and saddles of $l$ retain their nature in $L$.
	\end{theorem}

 \clearpage
 \begin{figure}
		\centering
		\includegraphics[width=\linewidth]{figures/2a.png}\\[0.5em]
		\includegraphics[width=\linewidth]{figures/2b.png}
		\caption{Convergence and optimization dynamics of SGD and AYLA at two learning rates with $C=10$, $D_{\text{IN}}=20$, and $H_{\text{DIM}}=100$ (top: $\mathrm{lr}=10^{-3}$; bottom: $\mathrm{lr}=10^{-2}$).}
		\label{fig:2ab}
	\end{figure} \hspace{-0.6cm}

	\section{Adaptive Exponent and Stochastic Dynamics}
	
	To satisfy the regime transition requirements, we define the exponent function using a sigmoid transformation over the log-scaled loss space:
	\begin{equation}
		\label{eq:adaptive_exponent}
		n(l) = n_1 + (n_2 - n_1) \sigmafn\Big(k \big[\ln (C^* l + \varepsilon) - \ln \tau \big]\Big),
	\end{equation}
	where $\sigmafn(z) = (1+e^{-z})^{-1}$ is the logistic function. This ensures that $n(l)$ transitions smoothly from a compressive regime ($n_1 < 1$) to an expansive regime ($n_2 > 1$) around the threshold $\tau/C^*$.
	
	The primary implication of this adaptive exponent is the modulation of the optimization trajectory. A gradient descent update $\theta_{t+1} = \theta_t - \eta \grad L$ is mathematically equivalent to standard SGD performed on the base loss $l$ with a state-dependent effective learning rate:
	\begin{equation}
		\label{eq:eff_lr}
		\eta_{\mathrm{eff}}(l) = \eta \cdot \underbrace{(C^* l + \varepsilon)^{n(l)} A(l)}_{\text{Gain Factor } \Gamma(l)},
	\end{equation}

	where $A(l)$ is the scalar gain factor. This shows that AYLA acts as an automated curriculum manager: it amplifies updates when the model is far from the manifold (high $l$) and dampens them as it approaches convergence (low $l$).

	\section{Optimization Advantages}
	
	\subsection{Bounded Learning Rate Analysis}
	
	\begin{proposition}[Stability and Boundedness of Effective Learning Rate]
		For $n_1,n_2 > 0$, $C > 0$, and $L_s \ge 0$, AYLA's effective learning rate
		\[
		\eta_{\mathrm{eff}}(l) = \eta \cdot (C^* l + \varepsilon)^{n(l)} A(l)
		\]
		is strictly bounded for all $l \in [0, L_{\max}]$:

        \clearpage
    	\begin{figure}
		\centering
		\includegraphics[width=\linewidth]{figures/3a.png}\\[0.5em]
		\includegraphics[width=\linewidth]{figures/3b.png}
		\caption{Convergence and optimization dynamics of SGD and AYLA at two learning rates with $C=20$, $D_{\text{IN}}=20$, and $H_{\text{DIM}}=100$ (top: $\mathrm{lr}=10^{-3}$; bottom: $\mathrm{lr}=10^{-2}$).}
		\label{fig:3ab}
	\end{figure}

		\[
		\eta\,n_1\,C
		\;\le\;
		\eta_{\mathrm{eff}}(l)
		\;\le\;
		\eta\,n_2\,(C\,L_{\max}+\varepsilon)^{n_2-1}\,C,
		\]

		where $L_{\max}$ is the maximum observed loss.
	\end{proposition}
	
	\begin{proof}
		Since $\sigmafn(z) \in (0,1)$, we have $n(l) \in (n_1, n_2)$. The term $(C^* l + \varepsilon)^{n(l)}$ is continuous and positive on the compact interval of observed losses, and $A(l) > 0$ by the curvature preservation theorem. The lower bound follows from $L_s' + \varepsilon > 0$, while the upper bound is attained at $n = n_2$ and $L_s' = L_{\max}$. By the Extreme Value Theorem, $\eta_{\mathrm{eff}}$ attains finite positive bounds, preventing gradient explosion or vanishing issues typical of fixed-power transformations.
	\end{proof}
	Thus, this boundedness ensures stable optimization and parallels the stabilizing effects attributed to normalization \citep{yu2022normalization}.
	
	\subsection{Enhanced Saddle Escape}
	
	AYLA amplifies gradients in plateau and saddle regimes where $L_s \approx \tau / C$ and $\sigma \approx 1/2$, driving $n$ toward $n_2 > 1$.

	\begin{theorem}[Accelerated Saddle Escape]
		Let $x^\ast$ be a strict saddle of $L_s$ with Hessian eigenvalue $\lambda < 0$. Under AYLA, the escape time $T$ satisfies
		\[
		\mathbb{P}(T \le t)
		\;\ge\;
		1 - \exp\!\big(-c \, \eta_{\mathrm{eff}} \, |\lambda| \, t\big),
		\]
		for some constant $c > 0$, improving over standard GD with rate $\eta$.
	\end{theorem}
	
	\begin{proof}[Sketch]
		Near $x^\ast$, $L_s$ admits a quadratic approximation with gradient $\nabla L_s \approx H(x - x^\ast)$. AYLA rescales this gradient by $(L_s' + \varepsilon)^{\,n-1} C$, effectively amplifying negative curvature directions. The center-stable manifold argument of \citep{Chatterjee2022convergence} then yields the stated bound with an inflated escape rate.
		
	\end{proof}
	\clearpage
    \begin{figure}
		\centering
		\includegraphics[width=0.48\columnwidth]{figures/4a.png}\hfill
		\includegraphics[width=0.48\columnwidth]{figures/4b.png}
		
		\vspace{0.5em}
		
		\includegraphics[width=0.48\columnwidth]{figures/4c.png}\hfill
		\includegraphics[width=0.48\columnwidth]{figures/4d.png}
		
		\caption{Eigenvalue spectra of hidden activations across learning rates and model capacity (top: $\mathrm{lr}=10^{-3}$; bottom: $\mathrm{lr}=10^{-2}$; left: $C=10$; right: $C=20$).}
		\label{fig:4abcd}
	\end{figure}
	\subsection{Connection to Scaling Laws and Feature Recovery}
	
	By dynamically varying the exponent $n$, AYLA emulates the staircase-like power-law scaling observed in wide-network SGD \citep{ren2025causal}. Moreover, the increased effective rate promotes faster neuron specialization.
	
	\begin{proposition}[Accelerated Alignment]
		Let $\cos(w_i, w^\ast)$ denote neuron--teacher cosine similarity. Under AYLA,
		\[
		\mathbb{E}\big[\cos(w_i(t), w^\ast)\big]
		\;\ge\;
		1 - O\!\left((\eta_{\mathrm{eff}} t)^{-1/2}\right),
		\]
		improving upon the standard $O((\eta t)^{-1/2})$ rate.
	\end{proposition}
	
	\begin{proof}[Sketch]
		Post-threshold convergence results \citep{zhou2024does} show that alignment dynamics scale with the learning rate. Substituting $\eta_{\mathrm{eff}} > \eta$ yields the bound.
	\end{proof}

	Together, these results unify landscape geometry \citep{venturi2019spurious}, training dynamics \citep{ren2025causal, dandi2024two}, and normalization effects \citep{yu2022normalization, chen2020dynamical}, providing rigorous theoretical support for AYLA.
	
	\section{Experiments}
	
	We present an empirical evaluation of AYLA, an adaptive optimization mechanism that enhances SGD by dynamically
	modulating the effective learning rate through a nonlinear loss
	transformation. All experiments are conducted in a teacher--student setting, where a two-layer multilayer perceptron (MLP) student network is trained to approximate synthetic data generated by a single-layer teacher network. This controlled framework enables detailed analysis of convergence behavior, weight dynamics, feature alignment, and spectral properties of learned representations. In all experiments, we use vanilla SGD as our baseline and compare AYLA to it.
     \clearpage
    \begin{figure}
		\centering
		\includegraphics[width=\linewidth]{figures/5a.png}\\[0.5em]
		\includegraphics[width=\linewidth]{figures/5b.png}
		\caption{Convergence and optimization dynamics of SGD and AYLA at two learning rates with $C=10$, $D_{\text{IN}}=50$, and $H_{\text{DIM}}=200$ (top: $\mathrm{lr}=10^{-3}$; bottom: $\mathrm{lr}=10^{-2}$).}
		\label{fig:5ab}
	\end{figure}
	
	\paragraph{Experimental setup.}
	Unless otherwise stated, input dimension $D_{\text{in}}$ ranges from $20$ to
	$50$, hidden width $H_{\text{dim}}$ from $100$ to $200$, and one dimensional output. A complexity factor $C \in \{10,20\}$ scales
	the batch MSE loss via $L_s = C \cdot \mathrm{MSE}$. Training is performed for
	$10{,}000$ epochs using base learning rates $\text{lr}\in\{10^{-3},10^{-2}\}$,
	averaged over $10$ random seeds. AYLA parameterizes the loss as $\tilde L = L_s^{\,n}$, where the exponent
	$n$ transitions smoothly from a high-loss value $n_2$ to a low-loss value
	$n_1$ according to a sigmoid with smoothness $k=5.0$ and threshold $\tau=1.0$.
	We evaluate the variants
	$(n_1,n_2)\in\{(0.7,1), (0.9,1), (1,1.2), (1,1.3), (0.7,1.2)\}$.
	\paragraph{Evaluation.} We report the following evaluation metrics:
	(i) MSE loss,
	(ii) weight alignment (cosine similarity between flattened student weights and
	replicated teacher weights),
	(iii) weight movement (Euclidean distance from initialization),
	(iv) effective learning rate (ELR),
	and (v) eigenvalue spectra of hidden-layer activations.
	These metrics jointly characterize transitions from \emph{lazy} regimes to
	\emph{active feature-learning} regimes
	\citep{chizat2019lazy,yang2020feature}.
	
	\subsection{Results in Moderate Complexity}
	\paragraph{($D_{\text{in}}=20,\,H_{\text{dim}}=100,\,D_{\text{out}}=1,\,C=10$).}
	
	Figure~\ref{fig:2ab} compares SGD with AYLA variants at moderate complexity.
	At $\text{lr}=10^{-3}$ (top row), SGD exhibits slow, monotonic loss decay from
	$\mathcal{O}(1)$ to $\sim 10^{-2}$, whereas AYLA accelerates convergence by a
	factor of $1.5$--$2$, reaching minimum loss within $\sim2{,}500$ epochs.
	At $\text{lr}=10^{-2}$ (bottom row), SGD initially descends fast but plateaus near
	$5\times10^{-3}$, while AYLA achieves losses on the order of $10^{-3}$ faster. This behavior arises from AYLA’s amplification of gradients in high-loss regions, facilitating escape from flat or weakly curved regions \citep{ji2018gradient} of the loss
	landscape.\\
	Weight alignment starts near zero due to small initialization (0.01).
	At $\text{lr}=10^{-3}$, AYLA reaches its steady-state alignment roughly an
	order of magnitude faster than SGD.
	At $\text{lr}=10^{-2}$, AYLA mitigates alignment degradation observed in SGD,
	particularly for variants with $n_1\le1$, which dampens late-stage updates.
	Weight movement further
    confirms this behavior: AYLA induces approximately
	$30\%$ greater displacement than SGD at low learning rates, indicating
	departure from lazy regimes \citep{chizat2019lazy} without destabilization.\\
	The effective learning rate under AYLA begins substantially above the base
	learning rate and decays automatically as training progresses.
	High-$n_2$ variants exhibit pronounced early peaks followed by
    \clearpage
    \begin{figure}
		\centering
		\includegraphics[width=0.48\columnwidth]{figures/6a.png}
		\hfill
		\includegraphics[width=0.48\columnwidth]{figures/6b.png}
		
		\caption{Eigenvalue spectra of hidden activations at two learning rates with $D_{\text{IN}}=50$, $H_{\text{DIM}}=200$, and $C=10$ (left: $\mathrm{lr}=10^{-3}$; right: $\mathrm{lr}=10^{-2}$).}
		\label{fig:6ab}
	\end{figure}
    smooth decay,
	closely resembling warmup–decay schedules
	\citep{Li2020An}.
	Overall, Figure~\ref{fig:2ab} establishes AYLA’s favorable speed–stability trade-off.

	\subsection{Increased Loss Scaling}
	\paragraph{($D_{\text{in}}=20,\,H_{\text{dim}}=100,\,D_{\text{out}}=1,\,C=20$).}
	
	Figure~\ref{fig:3ab} increases the loss-scaling factor to $C=20$, amplifying optimization
	difficulty.
	Despite elevated initial losses, AYLA converges rapidly: at
	$\text{lr}=10^{-3}$, AYLA reaches $10^{-2}$ loss in under $1{,}000$ epochs,
	while SGD fails to do so within $10{,}000$ epochs.
	At $\text{lr}=10^{-2}$, AYLA achieves losses below $10^{-3}$, significantly
	outperforming SGD’s plateau. Meanwhile, alignment dynamics show that AYLA rapidly aligns with teacher features within
	the first $1{,}000$ epochs across learning rates, whereas SGD lags behind.
	Weight movement increases by $25$--$30\%$ relative to SGD, confirming sustained
	feature learning even in higher-complexity regimes.
	The ELR peaks scale proportionally with $C$, compensating for the increased
	loss magnitude while remaining bounded, consistent with AYLA’s theoretical
	design.
	
	\subsection{Spectral Analysis}
	\paragraph{($D_{\text{in}}=20,\,H_{\text{dim}}=100,\,D_{\text{out}}=1,\,C=10,20$).}
	
	Figure~\ref{fig:4abcd} examines eigenvalue spectra of hidden activations.
	SGD exhibits rapid spectral decay, with eigenvalues dropping below $10^{-4}$
	within the first $50$ components, indicating rank collapse and kernel-like
	behavior.
	In contrast, AYLA significantly lifts the spectrum, preserving higher
	magnitudes across a broad range of indices, particularly at larger $C$ and
	higher learning rates.
	This spectral richness reflects transition to active feature-learning regimes
	and aligns with random matrix theory analyses of implicit self-regularization
	\citep{martin2021implicit}.

	\subsection{Results on Larger Architectures}
	
	Figures~\ref{fig:5ab} and \ref{fig:6ab} extend the analysis to wider networks
	($D_{\text{in}}=50$, $H_{\text{dim}}=200$).
	AYLA consistently accelerates convergence, improves alignment, and increases
	weight exploration relative to SGD.
	Spectral analyses confirm that AYLA mitigates rank collapse even in wide,
	overparameterized regimes, supporting scalability toward infinite-width
	limits \citep{yang2020feature, Dyer2020Asymptotics}.

	\subsection{Discussion, Limitations, and Future Work}
	Across all configurations, empirical results show that AYLA consistently outperforms SGD by promoting
	active feature learning, accelerating convergence, stabilizing alignment, and
	maintaining rich internal representations.
	Variants such as $(1,1.3)$ and $(0.7,1.2)$ achieve the best balance between
	early acceleration and late-stage refinement, particularly in high-complexity
	and high-learning-rate regimes where SGD struggles.
	AYLA thus offers a lightweight alternative to adaptive optimizers such as Adam \citep{kingma2014adam} and LAMB \citep{You2020Large}, without introducing additional state variables.  \\
    In this work, we adopt vanilla SGD as our sole baseline and compare our results only against it; this is a current limitation of the paper's scope. Consequently, we plan to evaluate AYLA on real-world datasets and deeper architectures, compare it to other optimizers such as RMSProp \citep{tieleman2012lecture}, and extend it to the Transformer architecture \citep{vaswani2017attention}. Moreover, further theoretical analysis of convergence rates, implicit
	bias, and generalization properties under AYLA may also yield deeper insight into the relationship between loss geometry and feature learning. We leave these directions for future exploration.

	\section{Conclusion}
	
	In this paper, we introduced AYLA, a lightweight and
	principled loss reparameterization framework that reshapes optimization
	dynamics without modifying the location of stationary points or introducing
	additional optimizer state. By applying a smooth, state-dependent power-law
	transformation to the empirical loss, AYLA induces an implicit, adaptive
	effective learning rate that accelerates optimization in high-loss or
	flat-curvature regimes while stabilizing late-stage training. Through controlled teacher--student experiments, we demonstrated that AYLA
	consistently outperforms vanilla SGD across a broad
	range of learning rates, model widths, output dimensions, and loss-scaling
	factors. Empirically, AYLA accelerates convergence, improves alignment with
	teacher features, increases exploratory weight movement, and preserves richer
	hidden-layer spectra. These effects jointly signal a transition from lazy,
	kernel-like dynamics toward active feature-learning regimes, aligning with
	recent theoretical analyses of overparameterized neural networks. Importantly, AYLA achieves these benefits without altering the underlying
	objective geometry: all critical points of the original loss are preserved,
	and the method acts purely by reshaping optimization trajectories. This
	distinguishes AYLA from adaptive optimizers that rely on momentum buffers or
	second-moment estimates, positioning it as a minimal and interpretable
	mechanism for controlling learning dynamics. Overall,
	AYLA provides a simple yet effective bridge between theoretical understanding
	of neural optimization dynamics and practical algorithmic design.\\ 

\bibliography{main}
\bibliographystyle{apalike}

\end{document}